\documentclass[10pt,twocolumn,letterpaper]{article}
\usepackage[pagenumbers]{pwstyle}
\usepackage{colortbl}

\usepackage{microtype}

\usepackage{xspace}
\usepackage{booktabs}
\usepackage{multirow}
\usepackage{amsmath,amssymb}
\usepackage{pifont}
\usepackage{placeins}
\newcommand{\method}{PersistWorld\xspace}%
\newcommand{\writer}{Visibility-aware State Writer\xspace}
\newcommand{\evolver}{State Evolver\xspace}
\newcommand{\reader}{Pose-queried Occupancy Reader\xspace}
\newcommand{\Writer}{Writer\xspace}
\newcommand{\Evolver}{Evolver\xspace}
\newcommand{\Reader}{Reader\xspace}

\definecolor{pwblue}{rgb}{0.21,0.49,0.74}
\usepackage[pagebackref,breaklinks,colorlinks,allcolors=pwblue]{hyperref}
\providecommand{\occavg}{\ifdefined\cellcolor\cellcolor{gray!12}\fi}
\providecommand{\occours}{\ifdefined\cellcolor\cellcolor{pwblue!8}\fi}

\title{Write, Evolve, Read: Streaming Cameras into\\a Persistent World State for 4D Occupancy Forecasting}

\author{%
ZiAn Wang$^{1}$\quad Tianyi Wang$^{2}$\quad Zhihao Chen$^{1}$\quad Manni Cui$^{3}$\quad Junkai Chen$^{4}$\\
ZhiJin Chen$^{4}$\quad Ruiqi Liu$^{4}$\quad Junfei Wu$^{4}$\quad Huanyu Zhang$^{4}$\quad Shu Wu$^{4,\dagger}$\\[0.2em]
{\small $^{1}$Jilin University\qquad $^{2}$Shanghai Jiao Tong University\qquad $^{3}$Huazhong University of Science and Technology}\\
{\small $^{4}$Institute of Automation, Chinese Academy of Sciences\qquad $^{\dagger}$Corresponding author}
}

\begin{document}
\flushbottom
\twocolumn[{%
\renewcommand\twocolumn[1][]{#1}%
\maketitle
\vspace{-2.2em}
\begin{center}
  \centering
  \captionsetup{type=figure}
  \includegraphics[width=\textwidth]{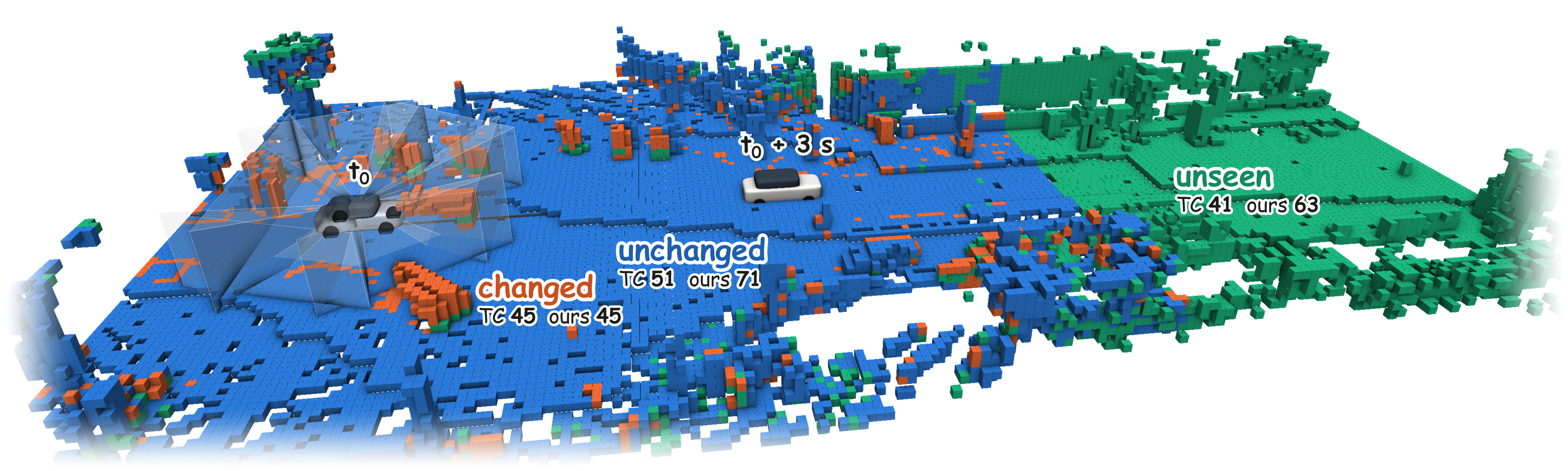}
  \vspace{-1.6em}
  \captionof{figure}{Ground-truth occupancy of an Occ3D-nuScenes scene 3\,s after the last input frame $t_0$.
  Blue voxels were seen in the input frames and are unchanged, orange voxels were seen but have changed, and green voxels were never seen.
    Under each label: IoU of SparseWorld-TC~\cite{du2026sparseworldtc} and ours, averaged over 1--3\,s.}
  \label{fig:teaser}
\end{center}%
\vspace{3pt}
}]
\begin{abstract}
4D occupancy forecasting predicts the 3D semantic occupancy of a driving scene seconds ahead.
Over a few seconds, the scene changes little while the vehicle moves, so most of the space ahead has already been seen.
Sorting the future into unchanged, changed, and unseen space, we find that current camera-based world models fall short on the unchanged part: they fail to hold on to what they have already seen, rather than to forecast what will change.
We therefore make what the model has seen its state. \method streams camera frames into a persistent 3D world state anchored in the scene, writing each frame once instead of rebuilding the state from past frames.
The state is learned end to end with the forecast, and an evolver predicts it at any queried time along the ego trajectory.
On Occ3D-nuScenes, \method raises the IoU of the best camera-based world model by 19 points at 1--3\,s and by 18 to 24 points up to 8\,s, while running more than twice as fast.
\par\vspace{3pt}{\upshape\footnotesize\raggedright\noindent Code: \url{https://github.com/wangzian828/PersistWorld}\\
Project: \url{https://wangzian828.github.io/PersistWorld}\par}
\end{abstract}

\section{Introduction}
\label{sec:intro}

\begin{figure}[t]
  \centering
  \includegraphics[width=\columnwidth]{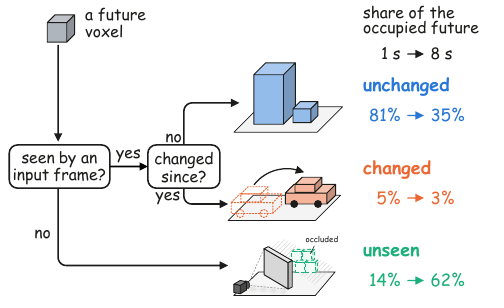}
  \caption{Two questions sort every occupied voxel of the future scene into unchanged, changed, or unseen space. Beside each kind: its share of the occupied future on Occ3D-nuScenes at 1 and 8\,s.}
  \label{fig:diagnosis}
\end{figure}

An autonomous vehicle must anticipate how the space around it will be occupied a few seconds ahead.
Occupancy world models forecast future 3D semantic occupancy from past observations and the planned ego trajectory, and much of the work on them has gone into the forecasting itself: how to tokenize the scene, how to generate its future, and how to condition it on the ego trajectory~\cite{zheng2024occworld,gu2024dome,shi2025come,liao2025i2world,du2026sparseworldtc}.
Yet with cameras, the setting that matters in practice, they remain far less accurate than models given ground-truth occupancy~\cite{liao2025i2world,gast2026,du2026sparseworldtc}.
Where do their errors lie?

Over a few seconds, a driving scene itself changes little, while the vehicle moves through it.
Most of the space ahead of the vehicle has therefore already been seen by its cameras.
We make this precise by sorting the future scene into three kinds of space (\cref{fig:teaser,fig:diagnosis}): \emph{unchanged} space, seen and still the same; \emph{changed} space, seen but different by then; and \emph{unseen} space, never seen.
One second ahead, four fifths of the occupied future is unchanged space; changed space stays small, and unseen space grows only as the vehicle moves into regions it could not see.

Current camera-based world models lose much of exactly this part.
SparseWorld-TC~\cite{du2026sparseworldtc}, the strongest of them, reaches an IoU of only 51.07 on unchanged space (\cref{fig:teaser}), where the scene as seen is already the answer.
Its own estimate of the current frame, moved to the target time without any forecasting, already reaches 48.68 there, and the occupancy that STCOcc~\cite{liao2025stcocc} predicts for two-stage methods reaches 46.31 (\cref{sec:analysis}).
Where forecasting is actually needed, on changed and unseen space, SparseWorld-TC is far ahead of carrying the seen scene forward, with an IoU of 45.35 and 40.92 against 26.33 and 0.
The shortfall of camera-based world models thus lies not in forecasting what will change but in holding on to what they have already seen, which models given ground-truth occupancy receive for free.

The same property of driving scenes also tells how to hold on to it: over a few seconds, the cameras view the same, largely unchanged scene from many positions.
We therefore make what the model has seen its state: a 3D world state anchored in the scene, written directly from camera images and learned end to end with the forecast.
The state stays in place as the vehicle moves, so the evidence that every frame provides about a place accumulates at one position.
To forecast, an evolver predicts the features of the state at the queried time along the ego trajectory.

Holding what has been seen in a state also makes forecasting cheap.
Since consecutive frames show nearly the same scene, the state is updated frame by frame as the images stream in: each new frame only adds its own evidence, as in streaming models that estimate the present scene~\cite{zuo2025gaussianworld,moon2026streamocc,chen2025gdfusion} or keep a memory for forecasting~\cite{mei2026irwm}.
Each image is thus processed once, and no separate perception network is needed, whereas sparse end-to-end models re-derive the scene from the image history for every forecast~\cite{dang2025sparseworld,du2026sparseworldtc} and two-stage methods first run a perception network~\cite{gu2024dome,shi2025come,liao2025i2world,gast2026}.
Since a forecast needs only a recent view of the surroundings, the state is reset periodically and keeps only the last few seconds of observation.

On Occ3D-nuScenes, our model raises the IoU and mIoU of the best camera-based world model by 19.13 and 3.32 at 1--3\,s and runs more than twice as fast.
Its IoU rises from 51.07 to 70.86 on unchanged space and from 40.92 to 62.71 on unseen space, and the same model leads by 18.19 to 24.41 IoU at every horizon up to 8\,s.
Our contributions are:
\begin{itemize}
  \item An analysis that sorts the future scene into unchanged, changed, and unseen space, and shows that camera-based world models fall short in holding on to what they have already seen rather than in forecasting what will change.
  \item A camera-based occupancy world model that holds what it has seen in a 3D world state anchored in the scene, written from images, learned end to end with the forecast, updated as a stream, and reset periodically.
  \item State-of-the-art camera-based forecasting on Occ3D-nuScenes at 1--3\,s and 1--8\,s with a single model, which is also the fastest of the compared methods.
\end{itemize}

\section{Related Work}
\label{sec:related}

\begin{figure*}[t]
  \centering
  \includegraphics[width=1.0\linewidth]{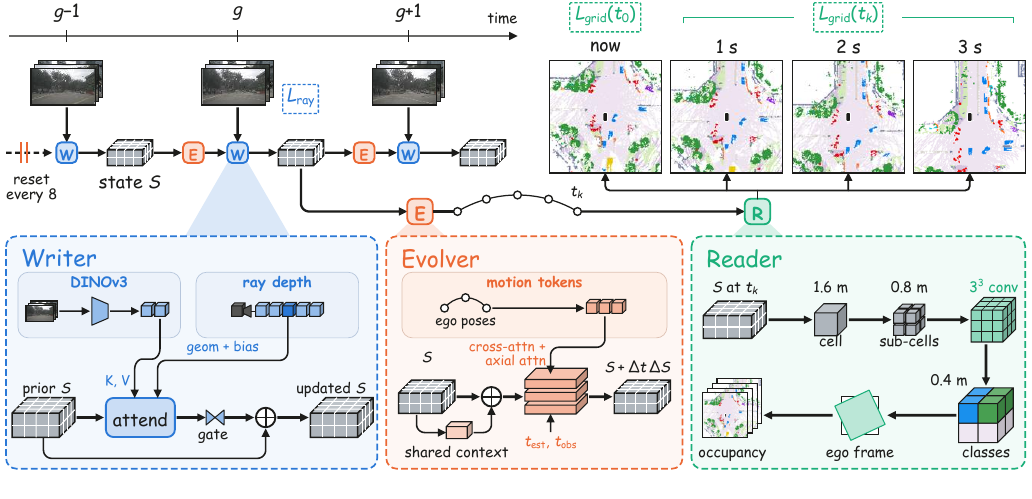}
  \caption{Overview of \method.
  Top: the world state $S$ is updated once per capture group; the \Evolver, marked E, carries it to the capture time of the next group, and the \Writer, marked W, writes the new images into it; $S$ is reset to the initial state every eight capture groups.
  To forecast, the \Evolver predicts the state at a queried time $t_k$ from $S$ and the ego trajectory, and the \Reader, marked R, decodes it in the queried ego frame.
  Bottom: the internal operations of each component, Secs.~\ref{sec:writer}--\ref{sec:reader}.
  The grid loss supervises the current and future occupancy, and the ray loss supervises the ray-depth sections of the \Writer.
  Images and ground-truth occupancy, seen from above, are from a nuScenes validation scene.}
  \label{fig:overview}
\end{figure*}

\paragraph{Occupancy from cameras.}
Camera-based occupancy perception lifts image features into a 3D grid of semantic voxels, through view transformation into bird's-eye-view or voxel features~\cite{li2022bevformer,li2023fbocc}, sparse queries~\cite{liu2024sparseocc,wang2024opus}, or 3D Gaussians~\cite{huang2024gaussianformer}, and is evaluated on benchmarks such as Occ3D~\cite{tian2023occ3d} and PanoBench~\cite{panobench2026}.
Rendering-based methods add supervision along camera rays, rendering depth and semantics from the predicted volume~\cite{pan2024renderocc,huang2024selfocc} or rendering both the prediction and the voxel labels and comparing them in the views or along the rays~\cite{chambon2025gaussrender,rayocc2026}.
We use the same kind of signal, predicting occupancy along each camera ray and supervising it with labels rendered from the ground-truth grid, so that the evidence of each image is written at the right depth along its rays.
Temporal methods carry information across frames by propagating object queries~\cite{wang2023streampetr} or fusing a long history through temporal stereo~\cite{park2023solofusion}; for occupancy, they evolve 3D Gaussians with ego and object motion~\cite{zuo2025gaussianworld}, stream voxel features~\cite{moon2026streamocc}, or unify several temporal cues~\cite{chen2025gdfusion}, and such a memory state can also be learned from multi-camera video as pre-training for perception and planning~\cite{min2024driveworld}.
All these methods estimate the present scene, and every new frame corrects their state.
Our state is also updated as a stream, but it is learned with the forecast and advanced to future times, where no new frame corrects it.

\paragraph{Forecasting the future scene.}
Occupancy world models tokenize the scene and generate its future autoregressively or by diffusion~\cite{zheng2024occworld,wei2024occllama,xu2025occllm,yan2024renderworld,zhang2025dfit,gu2024dome,jin2025occtens,xu2025dtt,cascadeocc2026,liao2025i2world}, warp the current scene to the future ego poses~\cite{gast2026}, add conditions~\cite{li2025uniscene,occwam2026,geniedrive2026}, or separate ego motion from scene change~\cite{shi2025come}.
They are commonly evaluated on Occ3D-nuScenes at 1, 2, and 3\,s~\cite{zheng2024occworld}, and up to 8\,s in later work~\cite{shi2025come,du2026sparseworldtc}.
These models take occupancy as input; with cameras, they forecast from a separately trained perception network~\cite{li2023fbocc,liao2025stcocc}, so what they forecast from is fixed before forecasting, and such chains degrade when errors enter upstream~\cite{occstress2026}.
Camera-based end-to-end models learn the forecast from images~\cite{ma2024cam4docc,yang2025driveoccworld,li2025preworld,forecastocc2026,geox2026}; some re-read the image history with sparse queries for every forecast~\cite{dang2025sparseworld,du2026sparseworldtc,sparseoccvla2026}, and some can be queried at any time~\cite{gem2026,agro2024uno}.
A persistent scene state has also been kept, but from bird's-eye-view features~\cite{mei2026irwm}.
\method holds what it has seen in a 3D world state written directly from images and updated as a stream, instead of re-deriving the scene for each forecast.
Which parts of the future must be forecast has also been studied: aggregating past observations is a strong baseline~\cite{khurana2023point}, most occupied future voxels are static~\cite{shi2025come}, and scene evolution has been split into static, moving, and newly observed parts~\cite{zuo2025gaussianworld,bai2027roadocc,ma2024cam4docc}.
We split the future by what the input frames have seen into unchanged, changed, and unseen space, and use the split to measure how each kind changes with the horizon and show that camera-based world models fall short on space they have already seen.

\section{Method}
\label{sec:method}

\noindent\textbf{Overview.}
\method holds what its cameras have seen in a world state $S$ on a fixed 3D lattice and updates it as a stream, once per capture group.
A capture group, which we also call a frame, is the surround-view images taken at one time, together with their camera parameters and the ego pose (Fig.~\ref{fig:overview}).
When a group arrives, the \evolver carries $S$ to its capture time (Sec.~\ref{sec:evolver}), the \writer writes the evidence of the new images into it (Sec.~\ref{sec:writer}), and the \reader decodes the current occupancy (Sec.~\ref{sec:reader}).
To forecast, the \Evolver predicts the state at each queried time from $S$ and the ego trajectory, and the \Reader decodes occupancy in the queried ego frame.
All components are trained end to end together with the image backbone (Secs.~\ref{sec:objective} and~\ref{sec:training}).

\subsection{World State}
\label{sec:state}
The state lives on a lattice of 1.6\,m cells anchored in the scene, so a cell keeps its address while the vehicle moves.
The region queried at any time is the occupancy volume around the ego, $80\times80\times6.4$\,m on Occ3D, and spans $50\times50\times4$ cells.
Each cell stores a feature that packs the features of its $2\times2\times2$ sub-cells of 0.8\,m, together with the time of its estimate and the time of its last observation, which condition the \Evolver (Sec.~\ref{sec:evolver}).
Packing keeps attention in the \Evolver over a compact lattice, while the \Writer and \Reader operate on the finer sub-cells; the \Reader decodes each sub-cell into $2\times2\times2$ voxels of 0.4\,m (Sec.~\ref{sec:reader}).
The state is stored only where it has been written; an unwritten cell takes a learned initial feature of its ego-relative position.

\subsection{\writer}
\label{sec:writer}
Images are encoded by a DINOv3 ViT-B/16~\cite{simeoni2025dinov3} whose last blocks are fine-tuned, and key and value maps are projected from its feature map.
Each capture group is encoded and written once, into the state that the \Evolver has just carried to its capture time.
For a sub-cell with center $x$, the \Writer samples keys and values bilinearly at the projection of $x$ into each camera, shifted by a small image-plane offset predicted from the sub-cell's feature, and adds a learned camera embedding.
Because the state is already at the capture time, the \Writer only places what the new images show and needs no motion model; all motion is left to the \Evolver.

\paragraph{Ray-depth sections.}
All sub-cells along a pixel ray project to the same image location and sample the same feature, so projection alone cannot tell them apart.
We split each patch ray into depth sections of 0.8\,m and let a light head predict, for every patch, a distribution over the section that holds the first occupied surface, with one more bin for rays that hit no surface.
For each sample, the section containing the sub-cell provides a geometry embedding that is added to the key and value, and the log-probability of that section is added as an attention bias, so evidence is weighted by how plausible it is that the sub-cell is the visible surface.

\paragraph{Update.}
The sub-cell feature attends, with a single query, to its samples from all cameras and to a null token with zero key and value, which lets it take no evidence when no camera shows it.
A gate from the prior and the candidate feature controls the residual update and is zero where no camera sees, so those keep the state carried by the \Evolver.

\subsection{\evolver}
\label{sec:evolver}
The \Evolver advances the state from the time of its estimate to a target time under the ego trajectory.
It carries the state to the capture time before writing and predicts the state at each queried future time.
Its inputs are the state in a window around the ego at the target time and the sequence of ego poses since the last reset (Sec.~\ref{sec:training}) up to the target pose.
Motion tokens are formed from the relative $\mathrm{SE}(3)$ transforms between consecutive poses and the time steps; each cell attends only to the motion that follows its own estimate, and is conditioned on the age of its estimate and of its last observation.
A context of the observed state, encoded by further axial blocks over all stored cells, is computed once per update of the state, shared by every advance from it, and added to the cell features before the residual blocks.
Residual blocks, with adaptive layer normalization conditioned on these ages, each apply cross-attention to the motion tokens and self-attention along one lattice axis.
The output is a change added to the state, scaled by the elapsed time, so that the \Evolver is the identity when no time passes.
Since the ego trajectory up to the target is an input, the same \Evolver serves every horizon.

\subsection{\reader}
\label{sec:reader}
The \Reader unpacks the sub-cells and applies a local residual block of a depth-wise $3^3$ convolution and point-wise mixing, followed by a linear layer that predicts class logits for the $2\times2\times2$ voxels of 0.4\,m inside each sub-cell; the logits are resampled into the queried ego frame.
The same \Reader decodes the current occupancy right after writing and the future occupancy from the predicted states.

\subsection{Unified Occupancy Objective}
\label{sec:objective}
\providecommand{\occavg}{\ifdefined\cellcolor\cellcolor{gray!12}\fi}
\providecommand{\occours}{\ifdefined\cellcolor\cellcolor{pwblue!8}\fi}
\providecommand{\OursSemOne}{\textbf{30.85}}
\providecommand{\OursSemTwo}{\textbf{28.59}}
\providecommand{\OursSemThree}{\textbf{26.72}}
\providecommand{\OursSemAvg}{\textbf{28.72}}
\providecommand{\OursGeoOne}{\textbf{68.43}}
\providecommand{\OursGeoTwo}{\textbf{67.88}}
\providecommand{\OursGeoThree}{\textbf{67.15}}
\providecommand{\OursGeoAvg}{\textbf{67.82}}
\providecommand{\OursFPS}{\textbf{23.15}}%
\begin{table*}[t]
\centering
\caption{Forecasting results on Occ3D-nuScenes~\cite{tian2023occ3d} at 1, 2, and 3\,s. Input: the input apart from the ego trajectory; -STC: forecasts from the occupancy predicted by STCOcc~\cite{liao2025stcocc}; Avg.: mean over 1--3\,s; FPS of two-stage methods includes their perception network; bold and underlined: best and second best camera-based result; --: not reported or not measured.}
\label{tab:main_occ3d}
\small
\setlength{\tabcolsep}{9.6pt}
\renewcommand{\arraystretch}{1.24}
\begin{tabular}{llccccccccc}
\toprule
\multirow{2}{*}{Method} & \multirow{2}{*}{Input} & \multicolumn{4}{c}{mIoU} & \multicolumn{4}{c}{IoU} & \multirow{2}{*}{FPS} \\
\cmidrule(lr){3-6} \cmidrule(lr){7-10}
 &  & 1\,s & 2\,s & 3\,s & \occavg Avg. & 1\,s & 2\,s & 3\,s & \occavg Avg. & \\
\midrule
OccWorld-O~\cite{zheng2024occworld} & Occ GT & 25.78 & 15.14 & 10.51 & \occavg 17.14 & 34.63 & 25.07 & 20.18 & \occavg 26.63 & -- \\
DOME~\cite{gu2024dome} & Occ GT & 35.11 & 25.89 & 20.29 & \occavg 27.10 & 43.99 & 35.36 & 29.74 & \occavg 36.36 & 1.04 \\
I$^2$-World~\cite{liao2025i2world} & Occ GT & 47.62 & 38.58 & 32.98 & \occavg 39.73 & 54.29 & 49.43 & 45.69 & \occavg 49.80 & 8.21 \\
GenieDrive~\cite{geniedrive2026} & Occ GT & 50.47 & 41.47 & 35.83 & \occavg 42.59 & 56.87 & 51.46 & 47.08 & \occavg 51.80 & 5.24 \\
CascadeOcc~\cite{cascadeocc2026} & Occ GT & 31.17 & 17.91 & 11.94 & \occavg 20.34 & 39.72 & 28.60 & 22.50 & \occavg 30.28 & -- \\
GAST~\cite{gast2026} & Occ GT & 52.61 & 44.53 & 38.84 & \occavg 45.33 & 59.37 & 54.90 & 50.99 & \occavg 55.09 & 17.41 \\
OccWAM~\cite{occwam2026} & Occ GT & 34.02 & 23.12 & 15.55 & \occavg 24.23 & 43.76 & 32.30 & 25.57 & \occavg 33.88 & -- \\
\midrule
OccWorld-D~\cite{zheng2024occworld} & Camera & 11.55 & 8.10 & 6.22 & \occavg 8.62 & 18.90 & 16.26 & 14.43 & \occavg 16.53 & -- \\
DOME-F~\cite{gu2024dome} & Camera & 24.12 & 17.41 & 13.24 & \occavg 18.25 & 35.18 & 27.90 & 23.44 & \occavg 28.84 & 0.99 \\
PreWorld~\cite{li2025preworld} & Camera & 12.27 & 9.24 & 7.15 & \occavg 9.55 & 23.62 & 21.62 & 19.63 & \occavg 21.62 & 0.95 \\
DTT~\cite{xu2025dtt} & Camera & 24.87 & 18.30 & 15.63 & \occavg 19.60 & 38.98 & 37.45 & 31.89 & \occavg 36.11 & -- \\
COME~\cite{shi2025come} & Camera & 26.56 & 21.73 & 18.49 & \occavg 22.26 & 48.08 & 43.84 & 40.28 & \occavg 44.07 & 0.25 \\
I$^2$-World-STC~\cite{liao2025i2world} & Camera & 28.15 & 23.62 & 20.08 & \occavg 23.95 & 43.93 & 40.59 & 37.37 & \occavg 40.63 & 2.87 \\
SparseWorld~\cite{dang2025sparseworld} & Camera & 14.93 & 13.15 & 11.51 & \occavg 13.20 & 22.96 & 22.10 & 21.05 & \occavg 22.03 & \underline{12.16} \\
SparseWorld-TC~\cite{du2026sparseworldtc} & Camera & 27.81 & \underline{25.31} & \underline{23.09} & \occavg \underline{25.40} & \underline{50.24} & \underline{48.82} & \underline{47.01} & \occavg \underline{48.69} & 9.36 \\
SparseOccVLA~\cite{sparseoccvla2026} & Camera & 15.86 & 13.54 & 11.72 & \occavg 13.71 & -- & -- & -- & \occavg -- & 5.47 \\
ForecastOcc~\cite{forecastocc2026} & Camera & 27.40 & 21.26 & 19.31 & \occavg 22.66 & 45.07 & 39.44 & 37.50 & \occavg 40.67 & 2.49 \\
GEM~\cite{gem2026} & Camera & 15.71 & 13.62 & 11.46 & \occavg 13.60 & 25.56 & 24.03 & 21.86 & \occavg 23.82 & -- \\
LiAuto-GeoX~\cite{geox2026} & Camera & 26.75 & 24.47 & 22.58 & \occavg 24.63 & 49.09 & 47.81 & 46.12 & \occavg 47.67 & -- \\
GAST-STC~\cite{gast2026} & Camera & \underline{29.22} & 24.42 & 20.46 & \occavg 24.70 & 45.91 & 42.56 & 38.94 & \occavg 42.47 & 3.48 \\
\occours \method (Ours) & \occours Camera & \occours \OursSemOne & \occours \OursSemTwo & \occours \OursSemThree & \occavg \OursSemAvg & \occours \OursGeoOne & \occours \OursGeoTwo & \occours \OursGeoThree & \occavg \OursGeoAvg & \occours \OursFPS \\
\bottomrule
\end{tabular}
\end{table*}

Each voxel is supervised with a focal score on the softmax probabilities $p_c$ of its class distribution, applied to every class as a binary event.
With the focal term \mbox{$f(q)=-(1-q)^2\log q$}, a voxel with label $y$ receives
\begin{equation}
\ell(p,y)=\omega_y\Big[a^{+}_{y}\,f(p_y)+\sum_{c\neq y}a^{-}_{c}\,f(1-p_c)\Big],
\label{eq:focal}
\end{equation}
where the class weights follow the class priors $\pi_c$ measured on the training set, $a^{+}_{c}\propto\pi_c^{-1/2}$ and $a^{-}_{c}\propto(1-\pi_c)^{-1/2}$ with a shared normalization constant, and $\omega_y{=}4$ for free voxels and $1$ otherwise.
Classes absent from a scene still receive the negative term, so rare classes are supervised in every scene.
The grid term $\mathcal{L}_{\text{grid}}(t)$ averages Eq.~\eqref{eq:focal} over the voxels inside the target mask at time $t$.

\paragraph{Grid and ray views.}
The ray-depth head of Sec.~\ref{sec:writer} predicts, for each patch ray, which section is occupied: it is a view of the same occupancy field along camera rays.
We render the per-section occupancy of the current frame from the ground-truth grid and supervise the ray view with the same focal objective.
The ray term $\mathcal{L}_{\text{ray}}$ averages Eq.~\eqref{eq:focal} over the valid sections, separately from the grid term, so the two views have equal weight.
The loss at a frame $t_0$ is
\begin{equation}
\mathcal{L}=\frac{1}{|\mathcal{K}|}\sum_{k\in\mathcal{K}}\mathcal{L}_{\text{grid}}(t_k)+0.25\big(\mathcal{L}_{\text{grid}}(t_0)+\mathcal{L}_{\text{ray}}\big),
\end{equation}
where $\mathcal{K}$ is the set of supervised future horizons at that frame.

\paragraph{Horizons.}
Each loss supervises 1, 2, and 3\,s, plus one horizon drawn uniformly from the available ones in 4--8\,s.
A single model therefore serves all horizons from 1 to 8\,s.

\subsection{Memory Span and Training}
\label{sec:training}
\providecommand{\occavg}{\ifdefined\cellcolor\cellcolor{gray!12}\fi}
\providecommand{\occours}{\ifdefined\cellcolor\cellcolor{pwblue!8}\fi}
\newcommand{\longours}[9]{\occours \method (Ours) & \occours Camera & \occours #1 & \occours #2 & \occours #3 & \occours #4 & \occours #5 & \occours #6 & \occours #7 & \occours #8 & \occavg #9 \\}
\begin{table*}[t]
\centering
\caption{Long-horizon forecasting results on Occ3D-nuScenes~\cite{tian2023occ3d} at 1 to 8\,s. Avg.: mean over 1--8\,s; bold: best camera-based result.}
\label{tab:long_occ3d}
\small
\setlength{\tabcolsep}{9.6pt}
\begin{tabular}{llccccccccc}
\toprule
Method & Input & 1\,s & 2\,s & 3\,s & 4\,s & 5\,s & 6\,s & 7\,s & 8\,s & \occavg Avg. \\
\midrule
\multicolumn{11}{c}{\textit{mIoU}} \\
\midrule
DOME~\cite{gu2024dome} & Occ GT & 30.10 & 21.35 & 17.36 & 14.86 & 12.61 & 11.03 & 10.00 & 9.34 & \occavg 15.83 \\
COME~\cite{shi2025come} & Occ GT & 33.78 & 24.57 & 21.35 & 18.25 & 15.84 & 13.85 & 12.99 & 11.96 & \occavg 19.07 \\
SparseWorld-TC~\cite{du2026sparseworldtc} & Camera & 27.81 & 25.31 & 23.09 & 21.06 & 18.45 & 15.90 & 13.67 & 11.90 & \occavg 19.65 \\
\longours{\textbf{30.85}}{\textbf{28.59}}{\textbf{26.72}}{\textbf{24.86}}{\textbf{22.93}}{\textbf{20.99}}{\textbf{19.03}}{\textbf{16.94}}{\textbf{23.86}}
\midrule
\multicolumn{11}{c}{\textit{IoU}} \\
\midrule
DOME~\cite{gu2024dome} & Occ GT & 39.04 & 31.20 & 27.14 & 24.73 & 22.32 & 20.28 & 19.05 & 17.97 & \occavg 25.21 \\
COME~\cite{shi2025come} & Occ GT & 44.20 & 36.25 & 32.86 & 30.03 & 26.93 & 24.70 & 23.30 & 21.44 & \occavg 29.96 \\
SparseWorld-TC~\cite{du2026sparseworldtc} & Camera & 50.24 & 48.82 & 47.01 & 45.43 & 42.70 & 39.67 & 36.92 & 34.37 & \occavg 43.15 \\
\longours{\textbf{68.43}}{\textbf{67.88}}{\textbf{67.15}}{\textbf{66.21}}{\textbf{64.87}}{\textbf{63.23}}{\textbf{61.33}}{\textbf{58.76}}{\textbf{64.73}}
\bottomrule
\end{tabular}
\end{table*}

\paragraph{Memory span.}
What a forecast needs from the past is a recent view of the surroundings, so the state holds only recent observations.
We reset it to the initial state at the start of each scene and after every eight capture groups, the history length used by SparseWorld-TC, so it holds between one and eight capture groups.
Each keyframe is answered with the state it holds at that moment.

\paragraph{Training on streams.}
Training runs the model as at inference.
A training segment is a run of up to eight consecutive capture groups of one scene, starting from the initial state; losses are placed at keyframes whose state depths, the number of capture groups written since the reset, are drawn uniformly from one to eight, so the model is trained at every depth it meets at inference.
Gradients are truncated to the last two writes before each loss: the earlier writes run without gradient, so memory does not grow with the segment length.
Each image is encoded once per segment.

\section{Experiments}
\label{sec:exp}

\subsection{Setup}
\label{sec:setup}

\paragraph{Dataset and metrics.}
Occ3D-nuScenes~\cite{tian2023occ3d} provides 700 training and 150 validation scenes with 18-class labels at 0.4\,m.
Following SparseWorld-TC~\cite{du2026sparseworldtc}, a forecast uses at most the last eight capture groups and the ground-truth future ego poses.
We forecast 1, 2, and 3\,s ahead from every validation keyframe, which gives 6019 anchor keyframes with 5719, 5419, and 5119 targets at 1, 2, and 3\,s.
We report the intersection over union of occupied voxels, IoU, and its mean over the semantic classes, mIoU, inside the camera mask of the target frame.
For long-horizon forecasting, we query the same model at every horizon from 1 to 8\,s.

\paragraph{Streaming evaluation.}
Our model processes each validation scene as a stream of keyframes, as it would run on a vehicle.
Each capture group is written into the state once, and each keyframe is answered with the state at that moment, which holds one to eight capture groups (\cref{sec:training}).
As in training, the state is reset at the start of each scene and after every eight capture groups.

\paragraph{Implementation details.}
\label{sec:impl}
The state is written at every keyframe, 0.5\,s apart.
We train for 20 epochs with AdamW at a learning rate of $3{\times}10^{-4}$; \cref{sec:suppl_impl} gives the remaining settings.

\paragraph{Baselines.}
Methods marked Occ GT forecast from the ground-truth occupancy, and camera-based methods forecast from images.
Two-stage methods marked -STC forecast from the occupancy that STCOcc~\cite{liao2025stcocc} predicts from the images.
We measure the speed in frames per second, FPS, for every method with released code and weights on one A800 with batch size one, from the inputs on the GPU to the occupancy of the current frame and of 1, 2, and 3\,s on the CPU.
For two-stage methods, the time includes their perception network.

\begin{figure}[t]
  \centering
  \includegraphics[width=1.0\linewidth]{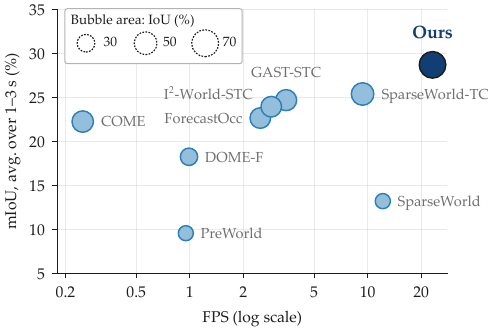}
  \caption{Speed and accuracy of forecasters on Occ3D-nuScenes. Bubble area: IoU averaged over 1--3\,s; FPS as in \cref{tab:main_occ3d}.}
  \label{fig:speed}
\end{figure}

\subsection{Main Results}
\label{sec:main_results}

\begin{figure*}[t]
  \centering
  \includegraphics[width=1.0\linewidth]{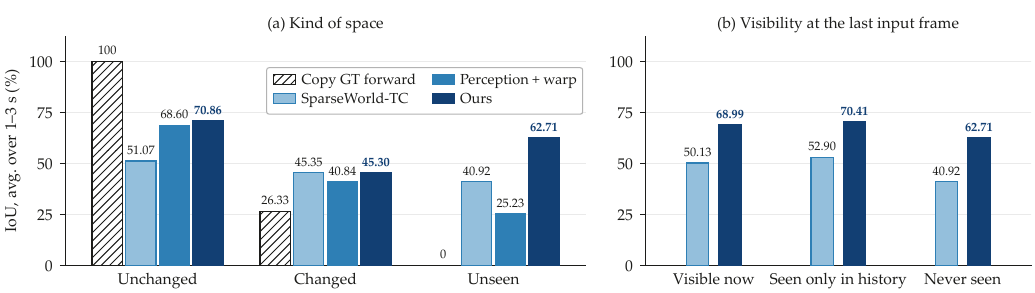}
  \caption{IoU of forecasts on Occ3D-nuScenes, averaged over 1--3\,s, split by kind of space in (a) and by visibility at the last input frame in (b). Copy GT forward: ground-truth occupancy of the input frames moved to the target frame with the ego poses; Perception + warp: our estimate of the current frame moved in the same way.}
  \label{fig:space_split}
\end{figure*}

\paragraph{Short-horizon forecasting.}
\cref{tab:main_occ3d} compares forecasts at 1, 2, and 3\,s.
Our model leads SparseWorld-TC, the strongest prior camera-based world model, at every horizon, and the lead grows with the horizon, from 18.19 to 20.14 IoU and from 3.04 to 3.63 mIoU.
With future ego poses planned by BEV-Planner~\cite{li2024egostatus} instead of the ground truth, the lead widens to 22.63 IoU and 4.00 mIoU, so it does not rest on the ground-truth future poses; \cref{sec:suppl_traj} gives the details.
Two-stage methods are limited by their input: with ground-truth occupancy, GAST leads I$^2$-World by 5.60 mIoU, but from the occupancy that STCOcc predicts, the two are within 0.75 mIoU and both fall below SparseWorld-TC (\cref{sec:analysis}).
Our model is also the fastest of all methods with a measured speed, at 23.15~FPS (\cref{fig:speed}), since it processes each image once and runs no separate perception network.

\paragraph{Long-horizon forecasting.}
\cref{tab:long_occ3d} queries the model of \cref{tab:main_occ3d} at every horizon from 1 to 8\,s.
It leads SparseWorld-TC by 18.19 to 24.41 IoU at every horizon, and by 21.58 IoU and 4.21 mIoU on average over 1--8\,s.
The lead holds at 8\,s, where most of the occupied future lies in space that no input frame saw (\cref{fig:diagnosis}), so the forecast does more than carry the seen scene forward: it also fills in unseen space.

\subsection{Three Kinds of Space}
\label{sec:analysis}

\begin{table}[t]
\centering
\small
\caption{Ablations of the world state on Occ3D-nuScenes. Each variant changes one property of the state and is trained with the same schedule. Scores are averaged over 1--3\,s; Unch., Chg., and Unseen: IoU on unchanged, changed, and unseen space (\cref{fig:diagnosis}); FPS as in \cref{tab:main_occ3d}.}
\label{tab:ablation}
\setlength{\tabcolsep}{3.6pt}
\renewcommand{\arraystretch}{1.08}
\newcommand{\ablatepending}{\textcolor{black!45}{--}}
\begin{tabular}{@{}lcccccc@{}}
\toprule
\multirow{2}{*}{Variant} & \multicolumn{2}{c}{Overall} & \multicolumn{3}{c}{IoU by space} & \multirow{2}{*}{FPS} \\
\cmidrule(lr){2-3} \cmidrule(lr){4-6}
 & mIoU & IoU & Unch. & Chg. & Unseen & \\
\midrule
\rowcolor{pwblue!8}
Full model & 28.72 & 67.82 & 70.86 & 45.30 & 62.71 & 23.15 \\
\midrule
Ego-centric state & 26.39 & 63.96 & 66.48 & 44.88 & 59.69 & 21.29 \\
Frozen encoder & 24.65 & 65.37 & 68.08 & 43.67 & 60.96 & 23.15 \\
No \Evolver & 24.78 & 62.25 & 67.20 & 42.87 & 50.49 & 37.52 \\
\midrule
Windowed & 29.26 & 67.97 & 71.47 & 46.82 & 61.51 & 12.02 \\
From STCOcc & 28.39 & 67.73 & 72.06 & 45.64 & 58.85 & 3.61 \\
\bottomrule
\end{tabular}
\end{table}

\paragraph{Where forecasts fall short.}
We split the occupied future by the two questions of \cref{sec:intro}, using the ground truth of the input frames.
Unchanged space makes up 80.85\% of it at 1\,s and 34.87\% at 8\,s, unseen space grows from 14.39\% to 62.06\%, and changed space stays below 5\% (\cref{fig:diagnosis}).
\cref{fig:space_split}a scores the forecasts at 1--3\,s on each kind.
As a reference, copying the ground truth of the input frames forward with the ego poses scores 100 IoU on unchanged space, 26.33 on changed space, and 0 on unseen space.
SparseWorld-TC does far better than this copy on changed and unseen space but reaches only 51.07 IoU on unchanged space, where the seen scene is already the answer.
Our model reaches 70.86 there and 62.71 on unseen space, 19.79 and 21.79 above it.

\paragraph{What is seen and what is forecast.}
To see what forecasting adds, we copy each model's estimate of the current frame forward in the same way; for our model, this is Perception + warp in \cref{fig:space_split}a.
On unchanged space, the copy already reaches 68.60 IoU for our model and 48.68 for SparseWorld-TC, close to their forecasts of 70.86 and 51.07, and the occupancy that STCOcc~\cite{liao2025stcocc} predicts for two-stage methods reaches 46.31.
Forecasting adds about as much there to both models, 2.26 and 2.39 IoU.
Accuracy on unchanged space is thus decided before forecasting, and our lead there comes from how well the model holds on to what it has seen.
On changed and unseen space, our forecast exceeds the copy by 4.46 and 37.48 IoU; this is what forecasting adds.

\paragraph{Visibility.}
The state also keeps what has left the view of the cameras.
\cref{fig:space_split}b splits the future by visibility at the last input frame.
On space seen only in earlier frames, our model reaches 70.41 IoU, slightly above the 68.99 on space still in view, so what has left the view is not lost.

\subsection{Ablation Study}
\label{sec:ablation}

\begin{figure}[t]
  \centering
  \includegraphics[width=\linewidth]{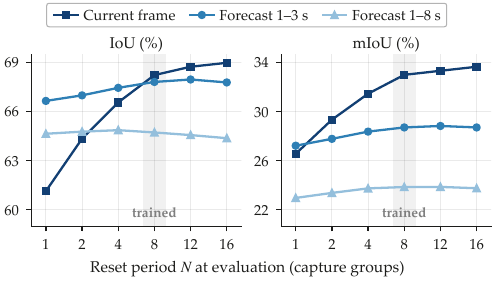}
  \caption{Effect of the reset period $N$ of the state at evaluation on Occ3D-nuScenes, for the model trained with $N{=}8$. Current frame: estimate of the current frame; forecasts averaged over 1--3\,s and 1--8\,s. Shaded band: the reset period used in training.}
  \label{fig:reset}
\end{figure}

\begin{figure*}[t]
  \centering
  \includegraphics[width=1.0\linewidth]{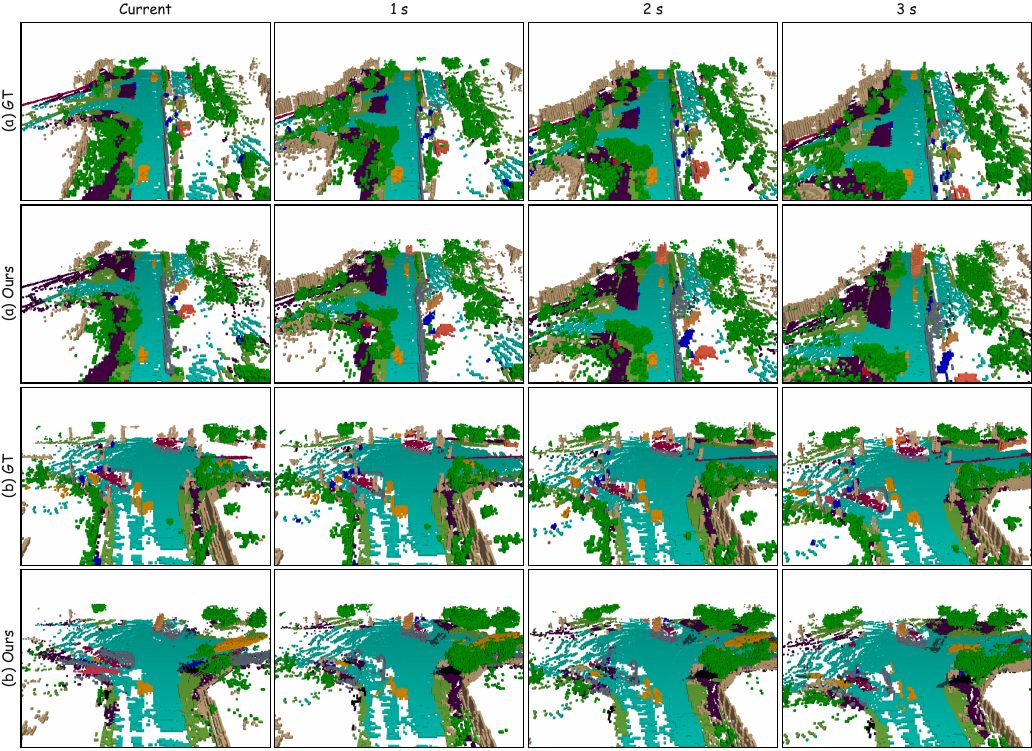}
  \caption{Forecasts of our model and the ground truth (GT) in two scenes (a) and (b) of Occ3D-nuScenes, at the current frame and 1, 2, and 3\,s ahead. Only voxels inside the camera mask of each target frame are drawn.}
  \label{fig:qual}
\end{figure*}

\paragraph{Properties of the state.}
\cref{tab:ablation} changes one property of the state at a time and scores each variant on the three kinds of space.
The state must be anchored in the scene: with an ego-centric state, whose lattice moves with the vehicle and is resampled into the ego frame of every new capture group, as in temporal fusion of bird's-eye-view features~\cite{li2022bevformer}, the IoU falls by 3.86, most on unchanged space, by 4.38, and by 3.02 on unseen space, against 0.42 on changed space.
Resampling the state at every capture group blurs what it has seen, and the extra resampling lowers the speed to 21.29~FPS.
The image features must be learned: with the image encoder frozen, the mIoU falls by 4.07 and the IoU on unchanged space by 2.78.
The \Evolver is what completes unseen space: without it, the state is read unchanged at each target pose, and the IoU on unseen space falls by 12.22, against 3.66 and 2.43 on unchanged and changed space.
Removing the depth conditioning or attention of the \Writer, or the shared context of the \Evolver, also lowers the accuracy of both the current frame and the forecasts, as the component ablations in \cref{sec:suppl_components} show.

\paragraph{Streaming.}
Updating the state as a stream doubles the speed at nearly the same accuracy.
Rebuilding the state for every query from the last eight capture groups reaches 29.26 mIoU and 67.97 IoU, against 28.72 and 67.82, but even with image features cached, every query then writes eight capture groups, and the speed nearly halves to 12.02~FPS.

\paragraph{Input.}
The lead does not come from our image encoder.
Writing the state from the occupancy that STCOcc~\cite{liao2025stcocc} predicts, instead of from images, reaches 28.39 mIoU and 67.73 IoU, while \mbox{GAST-STC} reaches 24.70 and 42.47 from the same input.
Copied forward, this input reaches only 46.31 IoU on unchanged space, and our state written from it reaches 72.06 there.
The state combines the predictions of successive frames at fixed places in the scene, so it holds more of the seen scene than any single prediction.
Running the perception network first, however, slows the model to 3.61~FPS, while a state learned end to end from images needs no separate perception network.

\paragraph{Memory span.}
A forecast needs only the last few seconds of observation.
We vary the reset period $N$ of the state at evaluation, for the model trained with $N{=}8$ (\cref{fig:reset}).
The estimate of the current frame improves with every added capture group, from 61.14 IoU at $N{=}1$ to 68.99 at $N{=}16$, as the evidence of successive frames accumulates in the state.
Forecasts at 1--3\,s level off at $N{=}8$ and change by at most 0.15 IoU beyond it, and the 1--8\,s average varies by at most 0.49.
Resetting the state every eight capture groups thus costs no forecasting accuracy.

\subsection{Qualitative Results}
\label{sec:qual}

\cref{fig:qual} shows forecasts of our model and the ground truth in two scenes, at the current frame and 1, 2, and 3\,s ahead.
In scene (a), the vehicle drives along a construction zone with moving traffic and pedestrians; the buildings, vegetation, and sidewalks along the road stay in place in our forecasts up to 3\,s, as in the ground truth, and the pedestrians on the right of the road remain in them.
In scene (b), the vehicle crosses an intersection, and our forecast continues the road into space that no input frame observed.
Comparisons with SparseWorld-TC on these and more scenes, forecasts from one state along two trajectories, and errors by distance and by dynamic and static classes are in \cref{sec:suppl_detail,sec:suppl_traj,sec:suppl_qual}.

\section{Conclusion}
\label{sec:conclusion}
Sorting the future scene into unchanged, changed, and unseen space shows that camera-based occupancy forecasting is limited first by how well models hold on to what they have already seen.
\method holds it in a 3D world state anchored in the scene and updated as a stream, and an evolver predicts the state at any queried time along the ego trajectory.
On Occ3D-nuScenes, this raises the IoU of the best camera-based world model by 19 points at 1--3\,s and keeps the lead up to 8\,s, at more than twice its speed.

{
    \small
    \bibliographystyle{ieeenat_fullname}
    \bibliography{main}
}
\appendix
\clearpage
\twocolumn[{\centering\Large\bf Appendix\par\vspace{1.2em}}]
\setcounter{dbltopnumber}{2}
\renewcommand{\dbltopfraction}{0.92}
\renewcommand{\textfraction}{0.1}
\setcounter{section}{0}
\renewcommand{\thesection}{\Alph{section}}

This appendix extends the experiments of \cref{sec:exp}.
\cref{sec:suppl_impl} gives implementation details, \cref{sec:suppl_inputs} writes the state from occupancy instead of images, \cref{sec:suppl_components} removes components of the model one at a time, \cref{sec:suppl_detail} breaks the forecasts down by class, distance, horizon, state depth, and ego motion, \cref{sec:suppl_traj} forecasts along trajectories other than the ground truth, and \cref{sec:suppl_qual} discusses more qualitative results.
All results use the protocol of \cref{sec:setup} unless stated otherwise.

\section{Implementation details}
\label{sec:suppl_impl}

Images are resized to $256{\times}704$ and, during training, augmented as in SparseWorld-TC~\cite{du2026sparseworldtc} by random scaling, cropping, flipping, and photometric distortion.
The last four blocks of DINOv3 are fine-tuned at one third of the base learning rate.
The state has 384 channels; the \Evolver has three residual blocks and three context blocks, and the \Writer two attention layers.
We train for 20 epochs with AdamW at a learning rate of $3{\times}10^{-4}$, decayed to one third over the last four epochs, with gradients clipped at a norm of 5, on six A800 GPUs in about 15 hours.

\section{Forecasting from occupancy}
\label{sec:suppl_inputs}

Our model can be written from occupancy instead of images.
For each new occupancy grid, the \Writer reads the labels of the eight voxels covered by each sub-cell, embeds their class histogram, and applies a gated update to the state; the \Evolver and \Reader are unchanged.
We train two such variants with the schedule of the main model.
Both use the same streaming protocol as the camera model: each new grid is written once into the state, which is reset every eight keyframes and at scene changes, so a forecast uses one to eight keyframes since the last reset.
\method-STC is written from the occupancy predicted by STCOcc~\cite{liao2025stcocc}, the perception network of the two-stage methods, and is scored inside the camera mask.
\method-GT is written from the ground-truth occupancy and is scored on the full volume, as the Occ GT methods in \cref{tab:main_occ3d}.
Its 1--3\,s evaluation uses 4519 anchors with at least 2\,s of history and 3\,s of future available in the scene; this selects the scored anchors and does not change the state reset schedule.
Each block of \cref{tab:suppl_occ} also moves the input of the current frame to the target frame with the ego poses, which forecasts as if nothing changed.

\begin{table*}[t]
\centering
\small
\caption{Forecasting from occupancy on Occ3D-nuScenes. + warp: the input occupancy of the current frame moved to the target frame with the ego poses; Avg.: mean over 1--3\,s. Rows with STCOcc input are scored inside the camera mask, rows with ground-truth input on the full volume. Bold and underlined: best and second best within each block, + warp rows excluded.}
\label{tab:suppl_occ}
\setlength{\tabcolsep}{13.5pt}
\begin{tabular}{lcccccccc}
\toprule
\multirow{2}{*}{Method} & \multicolumn{4}{c}{mIoU} & \multicolumn{4}{c}{IoU} \\
\cmidrule(lr){2-5} \cmidrule(lr){6-9}
 & 1\,s & 2\,s & 3\,s & \occavg Avg. & 1\,s & 2\,s & 3\,s & \occavg Avg. \\
\midrule
\multicolumn{9}{l}{\textit{Input: occupancy predicted by STCOcc}} \\
STCOcc + warp & 25.66 & 21.87 & 19.11 & \occavg 22.21 & 41.86 & 37.45 & 33.56 & \occavg 37.62 \\
GAST-STC~\cite{gast2026} & \underline{29.22} & \underline{24.42} & \underline{20.46} & \occavg \underline{24.70} & \underline{45.91} & \underline{42.56} & \underline{38.94} & \occavg \underline{42.47} \\
\occours \method-STC (Ours) & \occours \textbf{31.74} & \occours \textbf{28.07} & \occours \textbf{25.36} & \occavg \textbf{28.39} & \occours \textbf{69.45} & \occours \textbf{67.72} & \occours \textbf{66.02} & \occavg \textbf{67.73} \\
\midrule
\multicolumn{9}{l}{\textit{Input: ground-truth occupancy}} \\
GT + warp & 45.21 & 38.34 & 34.35 & \occavg 39.30 & 57.54 & 52.27 & 48.71 & \occavg 52.84 \\
OccWorld-O~\cite{zheng2024occworld} & 25.78 & 15.14 & 10.51 & \occavg 17.14 & 34.63 & 25.07 & 20.18 & \occavg 26.63 \\
DOME~\cite{gu2024dome} & 35.11 & 25.89 & 20.29 & \occavg 27.10 & 43.99 & 35.36 & 29.74 & \occavg 36.36 \\
I$^2$-World~\cite{liao2025i2world} & 47.62 & 38.58 & 32.98 & \occavg 39.73 & 54.29 & 49.43 & 45.69 & \occavg 49.80 \\
GenieDrive~\cite{geniedrive2026} & \underline{50.47} & 41.47 & 35.83 & \occavg \underline{42.59} & 56.87 & 51.46 & 47.08 & \occavg 51.80 \\
GAST~\cite{gast2026} & \textbf{52.61} & \textbf{44.53} & \textbf{38.84} & \occavg \textbf{45.33} & \underline{59.37} & \underline{54.90} & \underline{50.99} & \occavg \underline{55.09} \\
\occours \method-GT (Ours) & \occours 47.58 & \occours \underline{41.79} & \occours \underline{37.89} & \occavg 42.42 & \occours \textbf{64.98} & \occours \textbf{60.53} & \occours \textbf{56.79} & \occavg \textbf{60.77} \\
\bottomrule
\end{tabular}
\end{table*}

\paragraph{Occupancy-input forecasting.}
Using STCOcc predictions as the input source, \method-STC reaches 28.39 mIoU and 67.73 IoU on average, against 24.70 and 42.47 for GAST-STC, the strongest two-stage method.
From the ground truth, \method-GT reaches 42.42 mIoU and 60.77 IoU, against 45.33 and 55.09 for GAST: it is the most accurate in geometry and third in semantics.
\method-STC also exceeds the camera end-to-end baselines in \cref{tab:main_occ3d}, while reaching similar accuracy to our camera model.
Compared with copying only the current occupancy forward, the full model gains 30.11 IoU with STCOcc input and 7.93 with ground-truth input.

\paragraph{Same forecaster, different inputs.}
The three versions of our model share the \Evolver and \Reader and differ only in what is written into the state.
Writing from images reaches 28.72 mIoU and 67.82 IoU in \cref{tab:main_occ3d}, against 28.39 and 67.73 when writing STCOcc predictions with the same streaming update and reset schedule.
The state that the \Writer learns from images, trained end to end with the forecast, is thus as useful to the forecast as the output of a dedicated perception network trained on the same images.

\begin{table*}[t]
\centering
\small
\caption{Long-horizon forecasting from ground-truth occupancy on Occ3D-nuScenes at 1 to 8\,s, in the setting of COME~\cite{shi2025come}: one anchor per validation scene at its fourth keyframe, so the input spans 1.5\,s, with the ground-truth ego trajectory, scored on the full volume. + warp: the input occupancy of the current frame moved to the target frame with the ego poses; Avg.: mean over 1--8\,s; bold and underlined: best and second best within each block, + warp rows excluded.}
\label{tab:suppl_occ_long}
\setlength{\tabcolsep}{11.37pt}
\begin{tabular}{lccccccccc}
\toprule
Method & 1\,s & 2\,s & 3\,s & 4\,s & 5\,s & 6\,s & 7\,s & 8\,s & \occavg Avg. \\
\midrule
\multicolumn{10}{c}{\textit{mIoU}} \\
\midrule
GT + warp & 45.96 & 38.45 & 34.31 & 31.38 & 28.27 & 26.15 & 23.80 & 21.58 & \occavg 31.24 \\
DOME~\cite{gu2024dome} & 30.10 & 21.35 & 17.36 & 14.86 & 12.61 & 11.03 & 10.00 & 9.34 & \occavg 15.83 \\
COME~\cite{shi2025come} & 33.78 & 24.57 & 21.35 & 18.25 & 15.84 & 13.85 & 12.99 & 11.96 & \occavg 19.07 \\
GAST~\cite{gast2026} & \underline{36.23} & \underline{30.61} & \underline{27.13} & \underline{24.36} & \underline{21.98} & \underline{19.81} & \underline{17.83} & \underline{15.96} & \occavg \underline{24.12} \\
\occours \method-GT (Ours) & \occours \textbf{48.50} & \occours \textbf{42.37} & \occours \textbf{38.30} & \occours \textbf{34.96} & \occours \textbf{31.63} & \occours \textbf{29.07} & \occours \textbf{26.53} & \occours \textbf{24.12} & \occavg \textbf{34.43} \\
\midrule
\multicolumn{10}{c}{\textit{IoU}} \\
\midrule
GT + warp & 58.67 & 52.38 & 48.02 & 44.48 & 40.39 & 36.91 & 33.58 & 30.24 & \occavg 43.08 \\
DOME~\cite{gu2024dome} & 39.04 & 31.20 & 27.14 & 24.73 & 22.32 & 20.28 & 19.05 & 17.97 & \occavg 25.21 \\
COME~\cite{shi2025come} & 44.20 & 36.25 & 32.86 & 30.03 & 26.93 & 24.70 & 23.30 & 21.44 & \occavg 29.96 \\
GAST~\cite{gast2026} & \underline{49.09} & \underline{45.11} & \underline{42.11} & \underline{39.52} & \underline{37.19} & \underline{34.99} & \underline{32.87} & \underline{30.87} & \occavg \underline{39.06} \\
\occours \method-GT (Ours) & \occours \textbf{65.60} & \occours \textbf{60.64} & \occours \textbf{56.32} & \occours \textbf{52.36} & \occours \textbf{48.06} & \occours \textbf{44.58} & \occours \textbf{41.36} & \occours \textbf{37.60} & \occavg \textbf{50.82} \\
\bottomrule
\end{tabular}
\end{table*}

\paragraph{Up to 8\,s.}
\cref{tab:suppl_occ_long} queries \method-GT at every horizon from 1 to 8\,s in the long-horizon setting of COME, which GAST also follows.
\method-GT reaches 34.43 mIoU and 50.82 IoU on average over 1--8\,s, against 24.12 and 39.06 for GAST, and the warped current ground truth reaches 31.24 and 43.08.

\section{Component ablations}
\label{sec:suppl_components}

\begin{table}[t]
\centering
\small
\caption{Component ablations on Occ3D-nuScenes. Each variant removes one component and is trained with the schedule of the main model. Current: estimate of the current frame; 1--3\,s and 1--8\,s: forecasts averaged over these horizons.}
\label{tab:suppl_components}
\setlength{\tabcolsep}{2.6pt}
\begin{tabular}{@{}lcccccc@{}}
\toprule
\multirow{2}{*}{Variant} & \multicolumn{2}{c}{Current} & \multicolumn{2}{c}{1--3\,s} & \multicolumn{2}{c}{1--8\,s} \\
\cmidrule(lr){2-3} \cmidrule(lr){4-5} \cmidrule(lr){6-7}
 & mIoU & IoU & mIoU & IoU & mIoU & IoU \\
\midrule
\rowcolor{pwblue!8}
Full model & 33.01 & 68.24 & 28.72 & 67.82 & 23.86 & 64.73 \\
\midrule
No depth conditioning & 31.35 & 66.94 & 27.95 & 67.12 & 23.11 & 63.89 \\
Mean writing & 31.93 & 67.58 & 28.27 & 67.27 & 23.57 & 63.84 \\
No shared context & 32.66 & 67.97 & 28.31 & 67.53 & 23.68 & 64.50 \\
\bottomrule
\end{tabular}
\end{table}

\cref{tab:suppl_components} removes one component of the model at a time; each variant is trained with the schedule of the main model.
Without depth conditioning, the ray-depth head is still trained with the ray term, but neither its section embedding nor its log-probability enters the \Writer.
Mean writing averages the samples of each sub-cell instead of attending to them, so neither the learned query nor the depth bias weights the evidence.
Without the shared context, the \Evolver advances the state from the window around the ego alone.
Every component improves both the estimate of the current frame and the forecasts.
The ray-depth sections matter most for the current frame, by 1.66 mIoU and 1.30 IoU, as they decide at which depth along a ray the evidence of an image is written.

\section{Detailed results}
\label{sec:suppl_detail}

\paragraph{Per class.}
\cref{tab:suppl_perclass} lists the IoU of every class.
Our model gains most on terrain, manmade, drivable surface, sidewalk, and pedestrian, by 12.20 to 5.90 IoU, and SparseWorld-TC remains ahead on construction vehicle and barrier, by 3.97 and 1.40 IoU.
The large static classes, drivable surface, sidewalk, terrain, manmade, and vegetation, make up most of the occupied space, and our model is ahead on all five of them.

\newcommand{\pcls}[1]{\rotatebox{90}{#1}}
\begin{table*}[t]
\centering
\small
\caption{Per-class IoU on Occ3D-nuScenes, averaged over 1--3\,s.}
\label{tab:suppl_perclass}
\setlength{\tabcolsep}{3pt}
\resizebox{\linewidth}{!}{%
\begin{tabular}{lccccccccccccccccccc}
\toprule
Method & mIoU & IoU & \pcls{others} & \pcls{barrier} & \pcls{bicycle} & \pcls{bus} & \pcls{car} & \pcls{const.\ veh.} & \pcls{motorcycle} & \pcls{pedestrian} & \pcls{traffic cone} & \pcls{trailer} & \pcls{truck} & \pcls{drive.\ surf.} & \pcls{other flat} & \pcls{sidewalk} & \pcls{terrain} & \pcls{manmade} & \pcls{vegetation} \\
\midrule
SparseWorld-TC~\cite{du2026sparseworldtc} & 25.40 & 48.69 & 9.03 & 33.35 & 8.31 & 19.14 & 25.19 & 17.90 & 7.77 & 4.03 & 12.15 & 17.93 & 19.56 & 72.27 & 39.40 & 44.00 & 42.24 & 29.91 & 29.72 \\
\occours \method (Ours) & \occours 28.72 & \occours 67.82 & \occours 9.18 & \occours 31.95 & \occours 13.24 & \occours 20.27 & \occours 25.64 & \occours 13.93 & \occours 11.90 & \occours 9.93 & \occours 12.83 & \occours 22.99 & \occours 21.09 & \occours 79.50 & \occours 40.67 & \occours 50.78 & \occours 54.44 & \occours 37.92 & \occours 32.00 \\
\bottomrule
\end{tabular}}
\end{table*}

\begin{table}[t]
\centering
\small
\caption{Forecasts on Occ3D-nuScenes by distance from the ego vehicle and by class group, averaged over 1--3\,s. Dyn.: classes that can move; Static: all other classes.}
\label{tab:suppl_breakdown}
\setlength{\tabcolsep}{2.3pt}
\begin{tabular}{lcccccc}
\toprule
\multirow{2}{*}{Method} & \multicolumn{4}{c}{IoU by distance (m)} & \multicolumn{2}{c}{mIoU} \\
\cmidrule(lr){2-5} \cmidrule(lr){6-7}
 & 0--10 & 10--20 & 20--30 & 30--40 & Dyn. & Static \\
\midrule
SparseWorld-TC~\cite{du2026sparseworldtc} & 63.45 & 52.22 & 48.34 & 42.60 & 14.98 & 34.67 \\
\occours \method (Ours) & \occours 77.48 & \occours 67.17 & \occours 67.04 & \occours 66.68 & \occours 17.37 & \occours 38.81 \\
\bottomrule
\end{tabular}
\end{table}

\paragraph{By distance.}
\cref{tab:suppl_breakdown} splits the target voxels by their distance from the ego vehicle at the target time.
Both methods are most accurate within 10\,m.
Beyond it, SparseWorld-TC keeps losing accuracy with distance, from 52.22 to 42.60 IoU, while our model stays between 67.17 and 66.68 IoU, so our margin grows from 14.03 IoU within 10\,m to 24.08 IoU beyond 30\,m.
Far space is seen by few pixels in any single frame, and our state accumulates it over the input frames.

\paragraph{By class group.}
Dynamic classes are the vehicles, cyclists, and pedestrians that can move; the remaining classes are static.
On dynamic classes, our model reaches 17.37 mIoU, against 14.98 for SparseWorld-TC, and on static classes 38.81, against 34.67.
Dynamic objects are where most changed space lies, so this split complements the three kinds of space of \cref{fig:space_split}.

\begin{table}[t]
\centering
\small
\caption{Forecasts on Occ3D-nuScenes every 0.5\,s up to 8\,s. SparseWorld-TC: released checkpoint, evaluated by us; our whole-second rows are those of \cref{tab:long_occ3d}.}
\label{tab:suppl_halfsec}
\setlength{\tabcolsep}{6pt}
\begin{tabular}{ccccc}
\toprule
\multirow{2}{*}{Horizon} & \multicolumn{2}{c}{SparseWorld-TC} & \multicolumn{2}{c}{\method (Ours)} \\
\cmidrule(lr){2-3} \cmidrule(lr){4-5}
 & mIoU & IoU & \occours mIoU & \occours IoU \\
\midrule
0.5\,s & 29.00 & 50.19 & \occours 32.28 & \occours 68.62 \\
1.0\,s & 27.81 & 50.24 & \occours 30.85 & \occours 68.43 \\
1.5\,s & 25.92 & 48.51 & \occours 29.56 & \occours 68.13 \\
2.0\,s & 25.31 & 48.82 & \occours 28.59 & \occours 67.88 \\
2.5\,s & 23.35 & 46.32 & \occours 27.61 & \occours 67.49 \\
3.0\,s & 23.09 & 47.01 & \occours 26.72 & \occours 67.15 \\
3.5\,s & 21.95 & 46.35 & \occours 25.72 & \occours 66.61 \\
4.0\,s & 21.06 & 45.43 & \occours 24.86 & \occours 66.21 \\
4.5\,s & 19.98 & 44.30 & \occours 23.79 & \occours 65.37 \\
5.0\,s & 18.45 & 42.70 & \occours 22.93 & \occours 64.87 \\
5.5\,s & 17.19 & 41.29 & \occours 21.79 & \occours 63.70 \\
6.0\,s & 15.90 & 39.67 & \occours 20.99 & \occours 63.23 \\
6.5\,s & 14.78 & 38.32 & \occours 19.79 & \occours 61.74 \\
7.0\,s & 13.67 & 36.92 & \occours 19.03 & \occours 61.33 \\
7.5\,s & 12.66 & 35.54 & \occours 17.79 & \occours 59.26 \\
8.0\,s & 11.90 & 34.37 & \occours 16.94 & \occours 58.76 \\
\bottomrule
\end{tabular}
\end{table}

\paragraph{Every 0.5\,s.}
Because the \Evolver takes the ego trajectory up to the target time as input, the model can be queried at any time, including times it is never trained on.
Below 3\,s, the losses supervise only 1, 2, and 3\,s; \cref{tab:suppl_halfsec} also queries 0.5, 1.5, and 2.5\,s.
Our accuracy decreases steadily across all horizons, and every half-second horizon lies between its whole-second neighbors: from 1.5 to 7.5\,s, it deviates from their linear interpolation by at most 0.22 mIoU and 0.79 IoU.
For SparseWorld-TC, the IoU at 1.5 and 2.5\,s falls below both whole-second neighbors, 1.59 IoU under the interpolation at 2.5\,s.
Our forecast thus varies continuously with the queried time.

\begin{table}[t]
\centering
\small
\caption{Forecasts of \method on Occ3D-nuScenes by state depth, the number of capture groups written since the last reset when a keyframe is answered, averaged over 1--3\,s.}
\label{tab:suppl_depth}
\footnotesize
\setlength{\tabcolsep}{2.5pt}
\begin{tabular}{lcccccccc}
\toprule
Depth & 1 & 2 & 3 & 4 & 5 & 6 & 7 & 8 \\
\midrule
Current IoU & 61.20 & 67.70 & 68.53 & 69.06 & 69.53 & 69.81 & 70.04 & 70.29 \\
mIoU & 27.27 & 28.31 & 28.80 & 28.99 & 29.12 & 29.10 & 29.25 & 29.32 \\
IoU & 66.65 & 67.33 & 67.67 & 67.86 & 68.14 & 68.24 & 68.43 & 68.50 \\
\bottomrule
\end{tabular}
\end{table}

\paragraph{State depth.}
The state is reset after every eight capture groups, so a keyframe is answered with a state that holds between one and eight of them.
\cref{tab:suppl_depth} groups the keyframes by this number.
The estimate of the current frame improves quickly after a reset, from 61.20 IoU with one capture group to 67.70 with two and 70.29 with eight.
From three groups on, the forecast stays within 1 IoU of the 68.50 it reaches with eight.
The state thus recovers within about 1.5\,s after a reset, three capture groups at 0.5\,s spacing, which is why a short memory span costs little.

\begin{table}[t]
\centering
\small
\caption{Forecasts on Occ3D-nuScenes by ego motion over the forecast, averaged over 1--3\,s. Stopped: below 1\,m/s; Turn: heading change above 30 degrees within 3\,s; Straight: the remaining keyframes.}
\label{tab:suppl_motion}
\setlength{\tabcolsep}{2.7pt}
\begin{tabular}{lcccccc}
\toprule
\multirow{2}{*}{Method} & \multicolumn{2}{c}{Stopped} & \multicolumn{2}{c}{Straight} & \multicolumn{2}{c}{Turn} \\
\cmidrule(lr){2-3} \cmidrule(lr){4-5} \cmidrule(lr){6-7}
 & mIoU & IoU & mIoU & IoU & mIoU & IoU \\
\midrule
Share of keyframes & \multicolumn{2}{c}{18.7\%} & \multicolumn{2}{c}{75.7\%} & \multicolumn{2}{c}{5.7\%} \\
\midrule
SparseWorld-TC~\cite{du2026sparseworldtc} & 25.20 & 44.57 & 25.63 & 49.81 & 20.67 & 43.44 \\
\occours \method (Ours) & \occours 29.67 & \occours 64.07 & \occours 28.76 & \occours 68.88 & \occours 24.91 & \occours 63.24 \\
\bottomrule
\end{tabular}
\end{table}

\begin{figure*}[t]
  \centering
  \includegraphics[width=\linewidth]{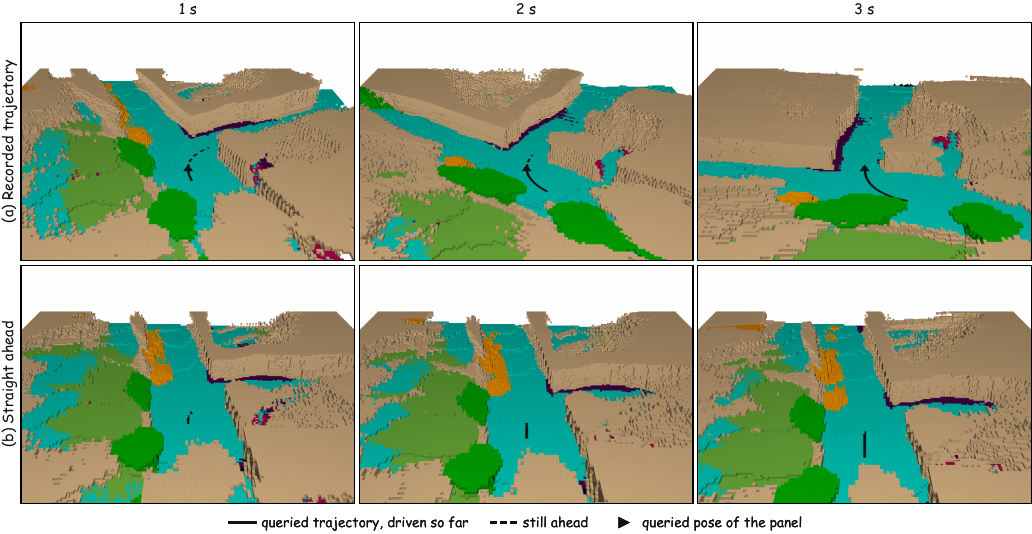}
  \caption{Forecasts of \method from one state along two trajectories at an intersection, at 1, 2, and 3\,s ahead: (a) the recorded right turn and (b) straight ahead. All forecast voxels are drawn, since a trajectory that was not driven has no camera mask.}
  \label{fig:suppl_traj}
\end{figure*}

\paragraph{Ego motion.}
How much of the target scene must be forecast depends on how the ego vehicle moves, and \cref{tab:suppl_motion} groups the keyframes accordingly.
In IoU, both methods are most accurate when the ego vehicle drives straight, less accurate when it stops, and least accurate in turns, where the target view swings into space that the input frames saw at grazing angles or not at all.
Our model reaches 68.88, 64.07, and 63.24 IoU in these three cases, and our margin over SparseWorld-TC is 19.07, 19.50, and 19.80 IoU.
The margin is thus nearly the same for every kind of ego motion.

\section{Trajectory conditioning}
\label{sec:suppl_traj}

Following the protocol of prior work~\cite{gu2024dome,liao2025i2world,shi2025come,du2026sparseworldtc}, \cref{tab:main_occ3d} gives every method the ground-truth future ego poses.
A deployed system would instead forecast along a planned trajectory, so we query both models with trajectories that are not the ground truth.
Constant velocity extrapolates the last observed ego motion; the planned trajectories are the BEV-Planner~\cite{li2024egostatus} predictions released by COME~\cite{shi2025come}, from the planner that SparseWorld-TC also uses.
SparseWorld-TC uses its released checkpoint with the same trajectories, so all its rows in \cref{tab:suppl_traj} come from our runs.
As in~\cite{du2026sparseworldtc}, a forecast made along a trajectory is scored against the ground truth of the frame the ego vehicle actually reaches, so a pose error counts fully against the forecast.

\begin{table}[t]
\centering
\small
\caption{Forecasts on Occ3D-nuScenes along different future ego trajectories, averaged over 1--3\,s. L2: error of the trajectory under the open-loop metric of BEV-Planner, averaged over 1--3\,s.}
\label{tab:suppl_traj}
\setlength{\tabcolsep}{2pt}
\begin{tabular}{lccccc}
\toprule
\multirow{2}{*}{Trajectory} & \multirow{2}{*}{L2 (m)} & \multicolumn{2}{c}{SparseWorld-TC} & \multicolumn{2}{c}{\method (Ours)} \\
\cmidrule(lr){3-4} \cmidrule(lr){5-6}
 & & mIoU & IoU & \occours mIoU & \occours IoU \\
\midrule
Ground truth & 0 & 25.40 & 48.69 & \occours 28.72 & \occours 67.82 \\
BEV-Planner & 0.48 & 18.37 & 38.79 & \occours 22.37 & \occours 61.42 \\
Constant velocity & 0.80 & 17.56 & 37.84 & \occours 21.34 & \occours 60.29 \\
\bottomrule
\end{tabular}
\end{table}

\begin{figure*}[t]
  \centering
  \includegraphics[width=0.85\linewidth]{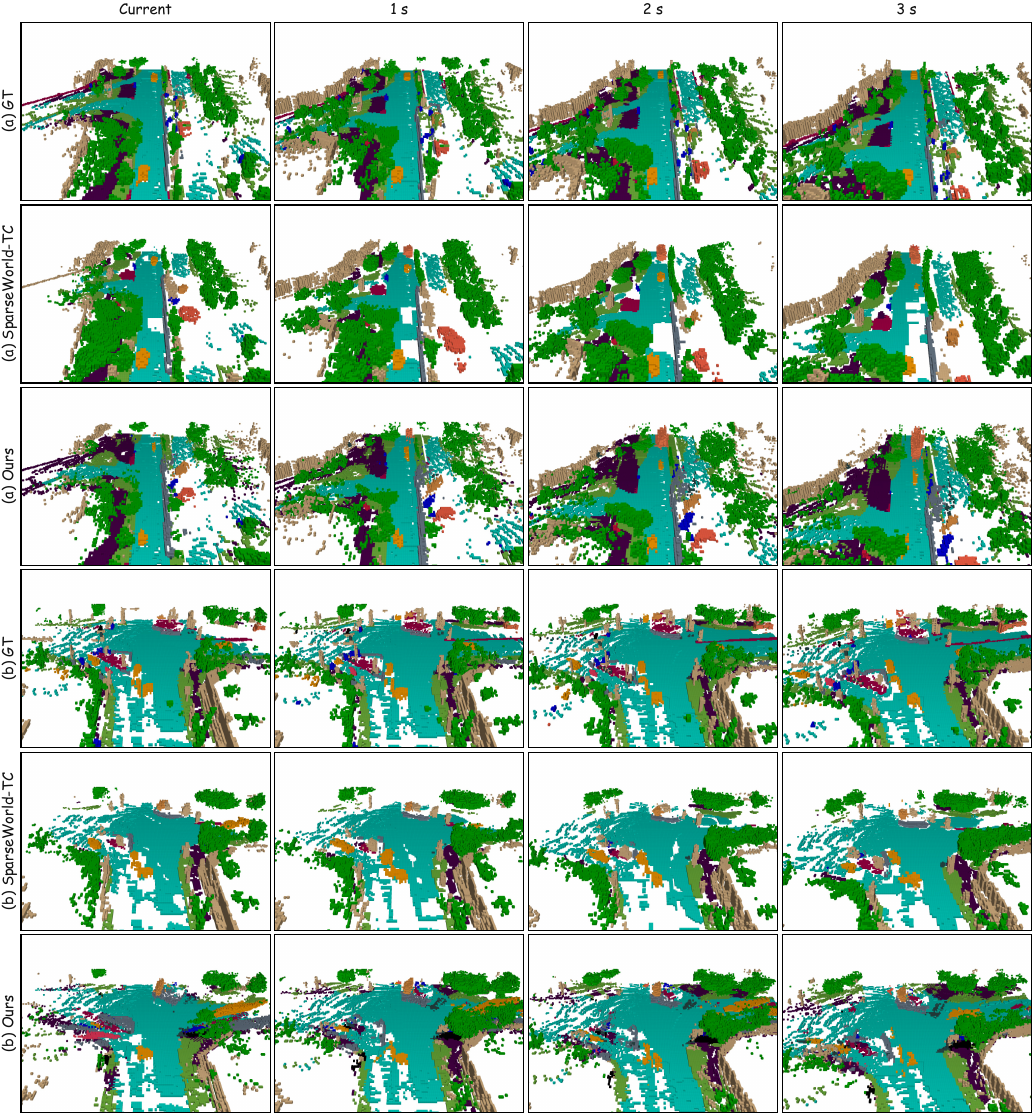}
  \caption{The scenes of \cref{fig:qual} with SparseWorld-TC, at the current frame and 1, 2, and 3\,s ahead: (a) moving traffic and pedestrians along a construction zone; (b) crossing an intersection into space that no input frame observed. Only voxels inside the camera mask of each target frame are drawn.}
  \label{fig:suppl_qual_main}
\end{figure*}

\begin{figure*}[t]
  \centering
  \includegraphics[width=0.85\linewidth]{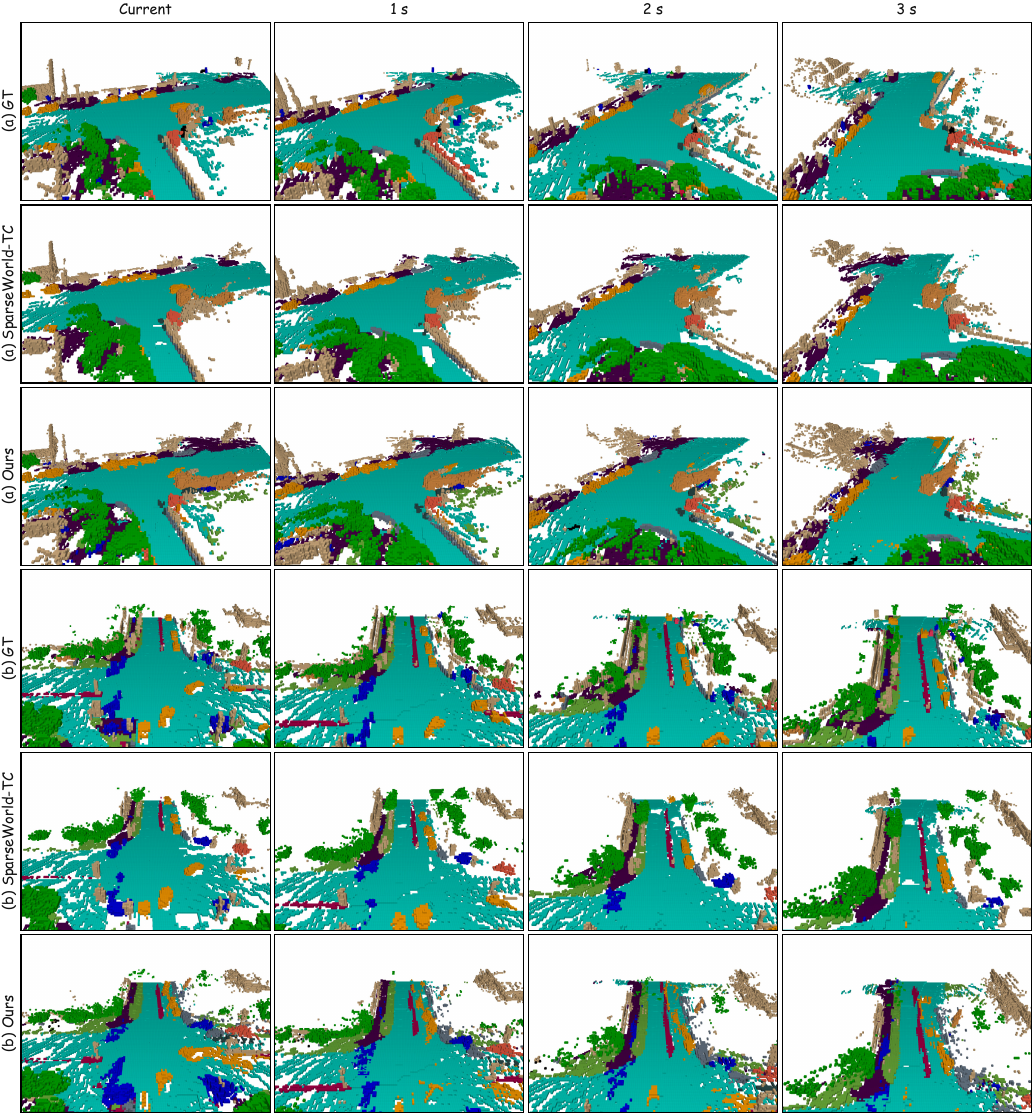}
  \caption{Two more scenes, drawn as in \cref{fig:suppl_qual_main}: (a) a right turn past cars parked along the curb; (b) pedestrians at an intersection.}
  \label{fig:suppl_qual}
\end{figure*}

\paragraph{Planned trajectories.}
\cref{tab:suppl_traj} shows that with the planner's trajectory, our model loses 6.35 mIoU and 6.40 IoU, and SparseWorld-TC loses 7.03 and 9.90.
With every trajectory source, our model stays ahead, by 4.00 mIoU and 22.63 IoU with the planner's trajectory and by 3.78 and 22.45 with constant velocity.

\paragraph{Different futures from one state.}
\cref{fig:suppl_traj} shows forecasts from the same observed state at an intersection along two trajectories: the recorded right turn and a straight path.
Along the recorded trajectory, the ego vehicle turns by about 70 degrees within 3\,s, and the forecast follows it into the cross street, which at 3\,s lies ahead between two blocks of buildings.
Straight ahead, the ego vehicle drives on toward the intersection, and the buildings, the trees, and the vehicle on the left of the road stay in place and come closer with each second.
The two rows thus show the same static structure from two sequences of poses, each forecast extending into the space that its trajectory reaches.

\section{More qualitative results}
\label{sec:suppl_qual}

\begin{figure*}[t]
  \centering
  \includegraphics[width=0.9\linewidth]{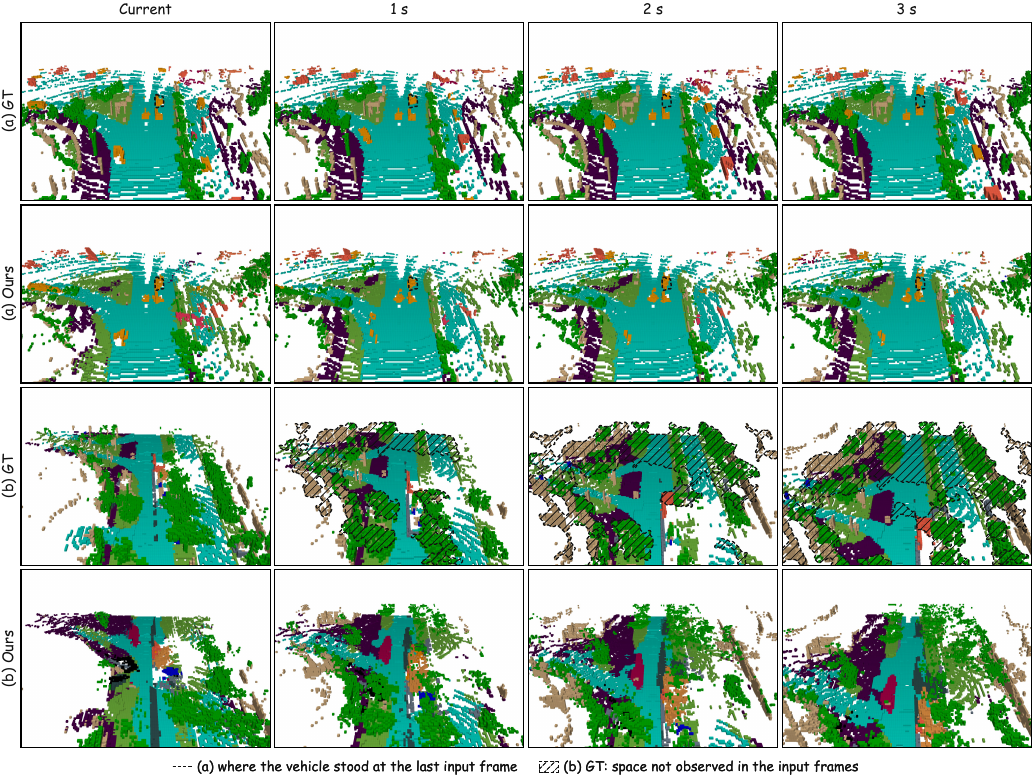}
  \caption{Forecasts of \method and the ground truth in two failure cases, at the current frame and 1, 2, and 3\,s ahead: (a) a car waiting ahead at an intersection drives off after the last input frame; (b) the ego vehicle drives into space that no input frame observed. Only voxels inside the camera mask of each target frame are drawn. Dashed outline in (a): where the car stood at the last input frame; hatching in (b): ground-truth space not observed in the input frames.}
  \label{fig:suppl_failure}
\end{figure*}

\cref{fig:suppl_qual_main} adds SparseWorld-TC to the two scenes of \cref{fig:qual}, and \cref{fig:suppl_qual} shows two more scenes chosen by the same rule.

\paragraph{Observed structure.}
In scene (a) of \cref{fig:suppl_qual_main}, the sidewalk on the left of the road stays in our forecasts up to 3\,s, while SparseWorld-TC loses most of it and leaves holes in the drivable surface.
In scene (a) of \cref{fig:suppl_qual}, the cars parked along the curb of the turn remain in place in our forecasts as the ego vehicle passes them.
Both are observed space that has not changed, which our state keeps.

\paragraph{Moving agents.}
In scene (a) of \cref{fig:suppl_qual_main}, our forecasts keep the pedestrians on the right of the road up to 3\,s, while SparseWorld-TC drops most of them after 1\,s.
In scene (b) of \cref{fig:suppl_qual}, both methods keep the pedestrians at the crossing, and our model stretches those walking along the left sidewalk into streaks at 2 and 3\,s.
Where an agent will be is uncertain, and our forecast spreads it over the places it may reach.

\paragraph{Unseen space.}
In scene (b) of \cref{fig:suppl_qual_main}, the ego vehicle enters an intersection whose far side no input frame observed.
SparseWorld-TC completes it mostly as drivable surface, while our model places vehicles along the cross street whose shapes are stretched and mixes classes near the corners.
Completing never-observed space remains the hardest part of the forecast for both methods.

\paragraph{Failure cases.}
In scene (a) of \cref{fig:suppl_failure}, the ego vehicle waits at an intersection, and a car about 15\,m ahead stands still in every input frame.
When the light turns green, the car drives off, and by 3\,s it has left its spot, but our forecast keeps a car there, filling 95\% of the spot at 3\,s.
The input frames hold no sign of the coming motion, so the state carries the car forward as static.
In scene (b), the ego vehicle drives about 30\,m in 3\,s, and most of the target scene lies in space that no input frame observed.
Of the occupied voxels revealed at 3\,s, our forecast fills about 60\% and gives 46\% the correct class: it continues the street, but the revealed buildings and trees come out thinner and more scattered than in the ground truth.
Such space can only be completed from what is typical of a street, not from the inputs.

\paragraph{Summary.}
Across these scenes, the structure that the input frames observed and that stays in place is kept by the state, while moving agents and newly revealed space are left to the \Evolver.
The errors arise where the input frames carry no evidence: a motion that has not yet started, and space that no camera has seen.

\end{document}